\documentclass[a4paper]{article}

\usepackage{INTERSPEECH2018}
\usepackage{multirow}
\usepackage{array}
\usepackage{wrapfig}
\usepackage{tabu}

\let\citet\newcite

\newcommand\uoi{\texttt{UoI}}

\usepackage{rotating}

\usepackage{tikz}
\usetikzlibrary{shapes.geometric}
\usetikzlibrary{patterns}
\usepackage{pgfplots}
\pgfplotsset{compat=1.14}

\title{A small Griko-Italian speech translation corpus}
\name{Marcely Zanon Boito$^{1}$, Antonios Anastasopoulos$^{2}$,\\ 
Marika Lekakou$^{3}$,
Aline Villavicencio$^{4,5}$, Laurent Besacier$^{1}$
}
\address{
  $^1$Laboratoire d'Informatique de Grenoble, Univ. Grenoble Alpes (UGA), France\\
  $^2$Department of Computer Science and Engineering, Univ. of Notre Dame, USA\\
  $^3$Department of Philology, Univ. of Ioannina, Greece\\
  $^4$ Institute of Informatics, UFRGS, Brazil\\
  $^5$ CSEE, University of Essex, UK
}
\email{contact: marcely.zanon-boito@univ-grenoble-alpes.fr and aanastas@nd.edu}

\begin{document}

\maketitle

\begin{abstract}

This paper presents an extension to a very low-resource parallel corpus collected in an endangered language, Griko, making it useful for computational research. The corpus consists of 330 utterances (about 20 minutes of speech) which have been transcribed and translated in Italian, with annotations for word-level speech-to-transcription and speech-to-translation alignments. The corpus also includes morphosyntactic tags and word-level glosses. Applying an automatic unit discovery method, pseudo-phones were also generated.
We detail how the corpus was collected, cleaned and processed, and we illustrate its use on zero-resource tasks by presenting some baseline results for the task of speech-to-translation alignment and unsupervised word discovery. The dataset is available online, aiming to encourage replicability and diversity in computational language documentation experiments.

\end{abstract}


\section{Introduction}

For many low-resource and endangered languages, speech data is easier to obtain than textual data. Oral tradition has historically been the main medium for passing cultural knowledge from one generation to the next, and at least~43\% of the world's languages are still unwritten~\cite{lewis2009ethnologue}. 
Traditionally, documentary records of endangered languages are created by highly trained linguists in the field. 
However, modern technology has the potential to enable creation of much larger-scale (but lower-quality) resources. Recently proposed frameworks~\cite{bird-EtAl:2014:Coling,stukerBulbCCURL2016} propose collection of bilingual audio, rendering the resource interpretable through translations.

New technologies have been developed to facilitate collection of spoken translations~\cite{bird-EtAl:2014:W14-22} along with speech in an endangered language, and there already exist recent examples of parallel speech collection efforts focused on endangered languages~\cite{blachon2016parallel, adda2016breaking,rialland2018}. The translation is usually in a high-resource language that functions as a \textit{lingua franca} of the area, as it is common for members of an endangered-language community to be bilingual.

Tackling the issue of the possible vanishing of more than 50\% of the current spoken languages by the year 2100~\cite{austin2011cambridge}, the Computational Language Documentation (CLD) field assembles two different communities: linguistics and informatics, proposing challenges~\cite{versteegh2016zero,jansen2013summary,zrc2017} and frameworks from speech signal~\cite{besacier2006towards,bartels2016toward,lignos2010recession}. However, as the interest on CLD approaches grows, it becomes clear the urgent need of more publicly available low-resource corpora to provide replicable evaluation of the proposed methods. We are aware of only a few endangered languages whose corpora are publicly available~\cite{godard2017very,hamlaoui2018}.

Our work is part of this effort to share resources, and with this paper we present a corpus on a truly endangered dialect from south Italy, Griko. The corpus has several levels of information (speech, machine extracted pseudo-phones, transcriptions, translations and sentence alignment), and we believe it can be an interesting resource for evaluating documentation techniques on (very) low-resource settings.

In addition, we provide baseline results for two tasks: speech-to-translation alignment and unsupervised word discovery.  
We encourage the community to challenge these results by using their own techniques.
For word discovery, we also provide the gold standard for evaluation following the Track 2 of the \textit{Zero Resource Challenge (ZRC) 2017} \cite{zrc2017}.
These metrics were extensively described in ~\cite{ludusan2014,zrc2017}, and are another important community effort for increasing reproducibility.

This paper is organized as follows: after a quick related work (section 2), the Griko language is presented (section 3). Data processing methodology (section 4) and dataset are then presented. Our baseline systems and results for two tasks (sections 5 and 6) are finally described.

\section{Related Work}\label{related}

\textbf{Unsupervised Word Discovery} (UWD) systems operate on unsegmented speech utterances and their goal is to 
output time-stamps delimiting stretches of speech, associated with class labels, corresponding to real words in the language.
This task is already considered in the Zero Resource Speech Challenge\footnote{http://zerospeech.com/2017} in a fully unsupervised setting: systems must learn to segment from a collection of raw speech signals only. Here, we investigate a slightly more favorable case where speech utterances are multilingually grounded (using cross-lingual supervision, where a written translation is available for each utterance).
In CLD scenarios, this task helps to attenuate the heavy charge on field linguists: the output vocabulary can be used as a first clue of the lexicon present in the language of interest.
As a monolingual setup, UWD was previously investigated from text input~\cite{Goldwater09bayesian} and from speech ~\cite{aren,lee2015unsupervised,bartels2016toward,elsner2013joint}.

The \textbf{speech translation} problem has been traditionally approached by feeding the output of a speech recognition system into a Machine Translation (MT) system. Speech recognition uncertainty was integrated with MT by using speech output lattices as input to translation models~\cite{ney1999speech,matusov2005integration}.
A sequence-to-sequence model for speech translation without transcriptions has been  introduced~\cite{duong2016attentional}, but was only evaluated on alignment. Synthesized speech data were translated in~\cite{berard2016listen} using a model similar to the Listen Attend and Spell model~\cite{chan2016listen}, while a larger-scale study~\cite{berard2018end} used an end-to-end system for translating audio books between French and English.
Sequence-to-sequence models to both transcribe Spanish speech and translate it in English have also been proposed~\cite{weiss2017sequence}, by jointly training the two tasks in a multitask scenario with two decoders sharing the speech encoder. This model was further extended \cite{anastasopoulos+chiang:naacl2018} with the translation decoder receiving information both from the speech encoder and the transcription decoder. 

For endangered languages (extremely low-resource settings) the lack of training data leads to the problem being framed as a sparse translation problem. 
This semi-supervised task lies between speech translation and keyword spotting, with cross-lingual supervision being used for word segmentation \cite{bansal2017towards,bansal2017weakly,anastasopoulos2017spoken,kamper2017visually}. Bilingual setups for word segmentation were discussed by~\cite{ip_stueker2008d,stuker:hal-00959225,StahlbergSVS12,boito2017unwritten}, but applied to speech transcripts (true phones). Among the most relevant to our approach are the works of \cite{duong2016attentional} on speech-to-translation alignment using attentional Neural Machine Translation (NMT) and of ~\cite{bansal2017weakly,anastasopoulos2017spoken} for language documentation. However, the former does not address word segmentation and is not applied to a language documentation scenario, while the latter methods do not provide a full coverage of the speech corpus analyzed. A neural approach for word segmentation in documentation scenarios using the soft attention matrices (which we also use for our baseline experiments) was investigated in \cite{boito2017unwritten}.

\section{The Griko Language}

Griko is a Greek dialect spoken in southern Italy, in the Grecìa Salentina area southeast of Lecce.\footnote{A discussion on the possible origins of Griko can be found in  \cite{manolessou2005greek}.} 
There is another endangered Italo-Greek variety in southern Italy spoken in the region of Calabria, known as Grecanico or Greco. Both languages, jointly referred to as \textit{Italiot Greek}, were included as seriously endangered in the UNESCO \textit{Red Book of Endangered Languages} in~1999.
Griko is only partially intelligible with modern Greek, and unlike other Greek dialects, it uses the Latin alphabet. In addition, it is rare among the Greek dialects, due to its retention of the infinitive in particular syntactic contexts. Less than~20,000 people (mostly people over 60 years old) are believed to be native speakers \cite{horrocks2009greek,douri2015griko}; unfortunately, this number is quite likely an overestimation \cite{chatzikyriakidis2010clitics}. 

Resources in Griko are very scarce, with almost no corpora  available for linguistic research. The first grammar of the language was composed by the German scholar Gerhard
Rohlfs \cite{rohlfs1977grammatica} to be followed by others \cite{karanastasis1997grammatiki}.

Recently, a corpus of Griko narratives was released \cite{anastasopoulos+al:coling2018}: it contains 114 narratives originally collected by Vito Domenico Palumbo (1854--1928) the most noted Griko scholar \cite{palumbo1998io,palumbo1999itela}. The narratives were further annotated with translations in Italian, and partly annotated with gold Part-of-Speech information.

Here, we present and extend the only Griko \emph{speech} corpus available online\footnote{\url{http://griko.project.uoi.gr}} \cite{grikodatabase}, consisting of about~20 minutes of speech in Griko, along with text translations into Italian. The original corpus (henceforth \uoi{} corpus, as it is hosted at the University of Ioannina, Greece) consists of~330 mostly elicited utterances by nine native speakers, annotated with transcriptions, morphosyntactic tags, and glossed in Italian. 

\section{Data Processing}

The original \uoi{} corpus was collected during a field trip in Puglia, Italy by two linguists, with a particular focus on the use of infinitive and verbal morphosyntax. 
The corpus contains utterances from~9 different speakers (5 male, 4 female) from the 4 villages (Calimera, Sternatia, Martano, Corigliano) where native speakers could still be found.
The digitally collected audio files (16-bit PCM, $44.1$kHz, stereo) were manually segmented into utterances, transcribed, glossed in Italian, and annotated with extensive morphosyntactic tags by a trained linguist. 

\subsection{Annotation Extensions}

In order to render the \uoi{} corpus useful for speech-related computational research on Griko, we extend the corpus with the following annotations:
\begin{enumerate}
    \item Free-form Italian translations for every utterance, created by a bilingual speaker,
    \item gold-standard word-level alignment information for every utterance, including annotated silences,
    \item gold-standard speech-to-translation alignments,
    \item pseudo-phones representation, obtained by using the acoustic unit discovery (AUD) method presented in~\cite{ondel2016variational},
    \item ZRC gold standard for standard evaluation, described in the next section.
\end{enumerate}

Figure~\ref{fig:example} shows an example of sentence pairs from our collection, and Table~\ref{table:stats} presents some statistics on these aligned transcriptions and translations. We observe that both sides of the parallel corpus are considerably similar with respect to the metrics presented here (sentence structure and vocabulary).
This is reasonable: the two languages belong to the same family and have been in contact for centuries.

\begin{figure*}[t]
      \centering
      \small
      \begin{tabular}{lcl} \midrule
      Griko & & \texttt{jat\`i \`iche polem\`isonta \`oli tin addom\`ada} \\
      Italian & & \texttt{perch\'e aveva lavorato tutta la settimana} \\
     \midrule
      \end{tabular}
      \caption{A tokenized and lower-cased sentence pair example in our Griko-Italian corpus.}
      \label{fig:example}
\end{figure*}
\begin{table*}[]
\centering
\begin{tabular}{l|cccccc} \toprule
\multicolumn{1}{c|}{} & \textbf{\# tokens} & \textbf{Vocabulary size} & \textbf{\begin{tabular}[c]{@{}c@{}}Average tokens\\ length\end{tabular}} & \textbf{\begin{tabular}[c]{@{}c@{}}Average \# tokens\\ per sentence\end{tabular}} & \textbf{\begin{tabular}[c]{@{}c@{}}Shortest\\ token\end{tabular}} & \textbf{\begin{tabular}[c]{@{}c@{}}Largest\\ token\end{tabular}} \\ \midrule
\textbf{Griko}        & 2,374              & 691                      & 5.68                                                                     & 7.19                                                                              & 1                                                                 & 16                                                               \\ \hline
\textbf{Italian}      & 2,384              & 456                      & 5.76                                                                     & 7.22                                                                              & 1                                                                 & 13             \\\bottomrule                                                 
\end{tabular}
\caption{Statistics of the 330 sentences in our parallel Griko-Italian corpus.}
\label{table:stats}
\end{table*}

\subsection{A reference compatible with the ZRC metrics}

In addition to the word-level annotations, we built and make available a reference (in the format of the ZRC challenge) in order to allow  evaluation of different word discovery approaches using this corpus. 
We had a manual alignment between speech and words, but no possibility to obtain  an accurate automatic alignment between speech and phones (or graphemes) due to the very small amount of data available (not possible to train an acoustic model using a Kaldi pipeline on 330 signals, for instance).

Thus, we used the word-level alignment information between speech and transcription, and the silence annotation available in our corpus, to approximate a speech-to-grapheme alignment. 
For each word present in the corpus, we retrieve its time window and segment this time window into smaller ones, giving 
to each existing grapheme an equal portion of its word time window.
We manually corrected some of the silence and word annotations to ensure that we had no overlap between silence and words time windows.
This approximation was necessary to make the ZRC metrics work. 

The final reference can be considered 
correct for evaluation of word discovery tasks (which do not take into account subword annotation), but should be consider with caution for evaluation of subword discovery tasks.
Finally, we created two ZRC versions, one removing the silence tokens, used for grapheme evaluation, and a second one with all the information, used for pseudo-phones evaluation.



\section{Speech-to-Translation Alignment}
\label{speechtotranslation}

The task of speech-to-translation alignment is the problem of identifying portions in an audio segment that should be aligned to words in (text) translation, without access to transcriptions \cite{duong2016attentional}. 
Our speech-to-translation alignment annotations allow us to evaluate such methods on our corpus. 
Evaluation is performed by computing standard precision, recall, and F-score on the links between speech frames and translation words.

Providing a baseline for future work, Table~\ref{tab:alignments} reiterates previous results on speech-to-translation alignment. We present results with three methods: a naive proportional baseline (\textbf{proportional}), a neural alignment model \cite{duong2016attentional} (\textbf{neural)}, and an unsupervised model (\textbf{DTW-EM}) \cite{anastasopoulos-chiang-duong:2016:EMNLP2016}.
The naive baseline assumes no reordering and simply segments the audio to as many segments as the translation words, each with a length proportional to the word's length in characters. The neural alignment model trains a speech-to-translation end-to-end sequence-to-sequence system with attention on \emph{all} the data, and then the soft attention matrices are converted to hard alignments between audio segments and translation words. \textbf{DTW-EM} is an unsupervised model that extends the IBM Model 2 alignment model \cite{och2001statistical} to work on speech segments, combining it with a Dynamic Time-Warping-based clustering approach \cite{petitjean2011global}.

Since the two languages have several similar characteristics, the naive proportional baseline is already very competitive; its recall is better than both other evaluated methods.
The unsupervised model, however, achieves much higher Precision and F-score than the rest.
Unsurprisingly, the neural approach performs significantly worse in this setting: $~330$ sentences are clearly not enough to train a robust word-level model.

\begin{table}[t]
\centering
\begin{tabular}{{@{}l|lll@{}}}
\toprule
\textbf{Method} & \textbf{P} & \textbf{R} & \textbf{F} \\ 
\midrule
\textbf{proportional} & 42.2 & \textbf{52.2} & 46.7 \\
\textbf{neural} & 24.6 & 30.0 & 27.0 \\
\textbf{DTW-EM} & \textbf{56.6} & 51.2 & \textbf{53.8} \\
\bottomrule
\end{tabular}
\caption{On speech-to-translation alignment, the unsupervised model outperforms the neural attentional model and the naive baseline in terms of Precision and F-score.}
\label{tab:alignments}
\end{table}

\section{Unsupervised Word Discovery Experiments}\label{worddiscovery}

\begin{table*}[h!]
\centering
\begin{tabular}{lllllllllllll} \toprule
                                            & \multicolumn{3}{c}{\textbf{dpseg}}                                                                & \multicolumn{3}{c}{\textbf{proportional}}                                                         & \multicolumn{3}{c}{\textbf{neural}}                                                               & \multicolumn{3}{c}{\textbf{merged neural}}                                                     \\ \midrule
\multicolumn{1}{l|}{}                       & \multicolumn{1}{c}{\textbf{P}} & \multicolumn{1}{c}{\textbf{R}} & \multicolumn{1}{c|}{\textbf{F}} & \multicolumn{1}{c}{\textbf{P}} & \multicolumn{1}{c}{\textbf{R}} & \multicolumn{1}{c|}{\textbf{F}} & \multicolumn{1}{c}{\textbf{P}} & \multicolumn{1}{c}{\textbf{R}} & \multicolumn{1}{c|}{\textbf{F}} & \multicolumn{1}{c}{\textbf{P}} & \multicolumn{1}{c}{\textbf{R}} & \multicolumn{1}{c}{\textbf{F}} \\ \hline
\multicolumn{1}{l|}{\textbf{grapheme}}      & \textbf{68.50}                          & \textbf{75.10}                          & \multicolumn{1}{l|}{\textbf{71.60}}      & 44.70                          & 44.80                          & \multicolumn{1}{l|}{44.70}      & 42.66                          & 51.84                          & \multicolumn{1}{l|}{46.72}      & 50.20                          & 54.00                          & 52.10    \\

 \hline
 \multicolumn{1}{l|}{\textbf{pseudo-phones}} & 23.30                          & \textbf{36.90}                          & \multicolumn{1}{l|}{28.50}      & 28.50                          & 29.90                          & \multicolumn{1}{l|}{29.20}      & 32.00                         & 27.68                          & \multicolumn{1}{l|}{29.56}      & \textbf{34.30}                          & 26.70                          & \textbf{30.00}                          \\

 \bottomrule                     
\end{tabular}
\caption{Boundary scores for the task of unsupervised word segmentation. Results for neural segmentation are the average over 5 runs. Best results for each metric are presented in bold.}
\label{table:UWS}
\end{table*}
\begin{table*}[]
\centering
\begin{tabular}{ccccccc} \toprule
                                               & \multicolumn{3}{c}{\textbf{grapheme}}                                                                                                                                                           & \multicolumn{3}{c}{\textbf{pseudo-phones}}                                                                                                                                                     \\ \hline
\multicolumn{1}{l|}{}                          & \multicolumn{1}{c}{\textbf{\# tokens}} & \multicolumn{1}{c}{\textbf{Vocabulary Size}} & \multicolumn{1}{c|}{\textbf{\begin{tabular}[c]{@{}c@{}}Average \# tokens \\ per sentence\end{tabular}}} & \multicolumn{1}{c}{\textbf{\# tokens}} & \multicolumn{1}{c}{\textbf{Vocabulary Size}} & \multicolumn{1}{c}{\textbf{\begin{tabular}[c]{@{}c@{}}Average \# tokens \\ per sentence\end{tabular}}} \\ \midrule
\multicolumn{1}{l|}{\textbf{proportional}}     & 2,370                                   & 1,715                                         & \multicolumn{1}{c|}{7.18}                                                                               & 2,366                                   & 1,431                                         & 7.17                                                                                                   \\
\multicolumn{1}{l|}{\textbf{dpseg}}            & 2,629                                   & 567                                          & \multicolumn{1}{c|}{7.97}                                                                               & 3,912                                   & 520                                          & 11.85                                                                                                  \\
\multicolumn{1}{l|}{\textbf{neural (average)}} & 2,972                                   & 1,462                                         & \multicolumn{1}{c|}{9.01}                                                                               & 1,929                                    & 1,066                                          & 5.84                                                                                                   \\
\multicolumn{1}{l|}{\textbf{merged neural}} & 2,573                                   & 1,476                                         & \multicolumn{1}{c|}{7.80}                                                                               & 1,676                                   & 967                                          & 5.08         \\ \bottomrule                                                                                        
\end{tabular}
\caption{A comparison between the generated segmentation by the four baselines. For the neural baseline, results are the arithmetic mean between the statistics for the 5 runs.}
\label{UWS:pseudographeme}
\end{table*}

In this section we illustrate the use of our corpus for the task of unsupervised word discovery. We use three different baselines, one monolingual and two bilingual, and two different representation levels, graphemes (from text) and pseudo-phones (automatically extracted from speech).
Evaluation is performed using the \textit{Boundary} metric from the \textit{Zero Resource Challenge 2017} (Track 2)~\cite{zrc2017}. 
We compute recall, precision and F-score. Below, we describe the three baselines evaluated in this work.

\begin{itemize}
\item \textbf{Dpseg (monolingual)}:
$dpseg$\footnote{Available at http://homepages.inf.ed.ac.uk/sgwater/resources.html.} is the non-parametric bayesian model introduced in \cite{goldwater2009bayesian}. On this setup, words are generated by a bigram model over a non-finity inventory, through the use of a Dirichlet-Process. Estimation is performed through Gibbs sampling. This approach is known as being very robust on low-resource scenarios. The hyper-parameters used here are the same from \cite{godard2016preliminary}.
\item \textbf{Proportional Segmentation (bilingual)}:
this baseline uses the word boundaries in the translation to segment the input \textit{proportionally}. We can expect considerable good results for proportional segmentation when applied on language pairs similar on sentence structure and average token length, and therefore, we expect good results for this baseline when applied to the Griko-Italian corpus (see Table~\ref{table:stats}).
\item \textbf{Neural Segmentation (bilingual)}: 
the method applied in this paper was presented in~\cite{boito2017unwritten}. It post-processes a NMT system's soft-alignment probability matrices to generate hard segmentation. Due to the length discrepancy between the symbols (graphemes and pseudo-phones) and the translations, our post-processing included alignment smoothing.
This procedure, proposed by \cite{duong2016attentional}, consists of 
adding temperature $T$ to the \textit{softmax} function used by the attention mechanism. Resulting soft-alignments matrices are further \textit{smoothed} by averaging each probability by its right and left neighborhood. However, in this work we use $T=1$ for all setups, and 
only the alignment matrices smoothing 
(averaging with the right and left neighbors) is used here.
Also, for stability reasons, we report the averaged scores over 5 different trained models.

\item \textbf{Merged Neural Segmentation (bilingual)}: the same methodology from the previous baseline, with the difference of averaging the soft-alignment probability matrices before post-processing, instead of averaging only the scores. We use the same 5 runs from the previous setup to generate an averaged (merged) segmentation. 
\end{itemize}

Table~\ref{table:UWS} presents the achieved results. Even on this very low-resource scenario, $dpseg$ has a remarkable performance for the task of word segmentation working with graphemes. It retrieved 75.10\% of the correct boundaries (recall). 
The second best method from the baselines for grapheme segmentation was the merged version of the neural segmentation.
The remaining two baselines (proportional and neural) had close performance, achieving retrieval between 44 and 52\%.


For the pseudo-phones segmentation, all methods had a considerable drop in performance, specially $dpseg$. They all achieved similar F-scores, with the merged neural baseline being slightly more effective.
Table~\ref{UWS:pseudographeme} presents some numbers for the generated segmentation of all methods presented in this section. We observe that, for pseudo-phones, $dpseg$ seems to over-segment the input (average tokens per sentence), while the neural baselines segmented the input considerably less.


Lastly, pseudo-phones were obtained through an unsupervised unit discovery system, which inevitably adds noise to the representation. This noise is then propagated to the word discovery system. 
We believe the achieved 
results for pseudo-phones illustrate the difficulty of the task of word discovery on extreme low-resource setups.

\section{Conclusion}
In this paper we presented an extension of a very small parallel corpus on an endangered language called Griko. We 
make this corpus, with all its different levels of representation, freely available to the community as an effort in the direction of research replicability for low-resource approaches.\footnote{Available at \url{goo.gl/EWa15G}}

We illustrated the potential of this parallel corpus by performing the tasks of speech-to-text alignment and unsupervised word discovery. We encourage the community to challenge the baselines presented here.

Future work includes comparing the tasks results from this extreme case of language documentation with other low-resource corpora, such as the one presented in~\cite{godard2017very}.

\section{Acknowledgements}

This work was partly funded by French ANR and German DFG under grant ANR-14-CE35-0002 (BULB project). Antonis Anastasopoulos was generously supported by NSF Award 1464553. Marika Lekakou was supported by the John S. Latsis Public Benefit Foundation under the project `Documentation and analysis of an endangered language: aspects of the grammar of Griko'.


\bibliographystyle{IEEEtran}
\bibliography{References}

\begin{thebibliography}{10}
\providecommand{\url}[1]{#1}
\csname url@samestyle\endcsname
\providecommand{\newblock}{\relax}
\providecommand{\bibinfo}[2]{#2}
\providecommand{\BIBentrySTDinterwordspacing}{\spaceskip=0pt\relax}
\providecommand{\BIBentryALTinterwordstretchfactor}{4}
\providecommand{\BIBentryALTinterwordspacing}{\spaceskip=\fontdimen2\font plus
\BIBentryALTinterwordstretchfactor\fontdimen3\font minus
  \fontdimen4\font\relax}
\providecommand{\BIBforeignlanguage}[2]{{%
\expandafter\ifx\csname l@#1\endcsname\relax
\typeout{** WARNING: IEEEtran.bst: No hyphenation pattern has been}%
\typeout{** loaded for the language `#1'. Using the pattern for}%
\typeout{** the default language instead.}%
\else
\language=\csname l@#1\endcsname
\fi
#2}}
\providecommand{\BIBdecl}{\relax}
\BIBdecl

\bibitem{lewis2009ethnologue}
M.~P. Lewis, G.~F. Simons, C.~D. Fennig \emph{et~al.}, \emph{Ethnologue:
  Languages of the world}.\hskip 1em plus 0.5em minus 0.4em\relax Dallas, TX:
  SIL International, 2009, vol.~16.

\bibitem{bird-EtAl:2014:Coling}
\BIBentryALTinterwordspacing
S.~Bird, L.~Gawne, K.~Gelbart, and I.~McAlister, ``Collecting bilingual audio
  in remote indigenous communities,'' in \emph{Proc. COLING}, 2014. [Online].
  Available: \url{http://www.aclweb.org/anthology/C14-1096}
\BIBentrySTDinterwordspacing

\bibitem{stukerBulbCCURL2016}
S.~St\"{u}ker, G.~Adda, M.~Adda-Decker, O.~Ambouroue, L.~Besacier, D.~Blachon,
  H.~Bonneau-Maynard, P.~Godard, F.~Hamlaoui, D.~Idiatov, G.-N. Kouarata,
  L.~Lamel, E.-M. Makasso, A.~Rialland, M.~Van~de Velde, F.~Yvon, and
  S.~Zerbian, ``Innovative technologies for under-resourced language
  documentation: The {Bulb} project,'' in \emph{Proceedings of CCURL
  (Collaboration and Computing for Under-Resourced Languages : toward an
  Alliance for Digital Language Diversity)}, Portoro\~{z} Slovenia, 2016.

\bibitem{bird-EtAl:2014:W14-22}
\BIBentryALTinterwordspacing
S.~Bird, F.~R. Hanke, O.~Adams, and H.~Lee, ``{Aikuma}: A mobile app for
  collaborative language documentation,'' in \emph{Proc. of the 2014 Workshop
  on the Use of Computational Methods in the Study of Endangered Languages},
  2014. [Online]. Available: \url{http://www.aclweb.org/anthology/W14-2201}
\BIBentrySTDinterwordspacing

\bibitem{blachon2016parallel}
\BIBentryALTinterwordspacing
D.~Blachon, E.~Gauthier, L.~Besacier, G.-N. Kouarata, M.~Adda-Decker, and
  A.~Rialland, ``Parallel speech collection for under-resourced language
  studies using the {LIG-Aikuma} mobile device app,'' in \emph{Proc. SLTU
  (Spoken Language Technologies for Under-Resourced Languages)}, vol.~81, 2016.
  [Online]. Available:
  \url{http://www.sciencedirect.com/science/article/pii/S1877050916300448}
\BIBentrySTDinterwordspacing

\bibitem{adda2016breaking}
G.~Adda, S.~St{\"u}ker, M.~Adda-Decker, O.~Ambouroue, L.~Besacier, D.~Blachon,
  H.~Bonneau-Maynard, P.~Godard, F.~Hamlaoui, D.~Idiatov \emph{et~al.},
  ``Breaking the unwritten language barrier: The bulb project,'' \emph{Procedia
  Computer Science}, vol.~81, pp. 8--14, 2016.

\bibitem{rialland2018}
A.~Rialland, M.~Adda-Decker, G.-N. Kouarata, G.~Adda, L.~Besacier, L.~Lamel,
  E.~Gauthier, P.~Godard, and J.~Cooper-Leavitt, ``{Parallel Corpora in Mboshi
  (Bantu C25, Congo-Brazzaville)},'' in \emph{LREC 2018 (in press)}, Japan,
  2018.

\bibitem{austin2011cambridge}
P.~K. Austin and J.~Sallabank, \emph{The Cambridge handbook of endangered
  languages}.\hskip 1em plus 0.5em minus 0.4em\relax Cambridge University
  Press, 2011.

\bibitem{versteegh2016zero}
M.~Versteegh, X.~Anguera, A.~Jansen, and E.~Dupoux, ``The zero resource speech
  challenge 2015: Proposed approaches and results,'' \emph{Procedia Computer
  Science}, vol.~81, pp. 67--72, 2016.

\bibitem{jansen2013summary}
A.~Jansen, E.~Dupoux, S.~Goldwater, M.~Johnson, S.~Khudanpur, K.~Church,
  N.~Feldman, H.~Hermansky, F.~Metze, R.~Rose \emph{et~al.}, ``A summary of the
  2012 jhu clsp workshop on zero resource speech technologies and models of
  early language acquisition,'' 2013.

\bibitem{zrc2017}
E.~Dunbar, X.~Nga~Cao, J.~Benjumea, J.~Karadayi, M.~Bernard, L.~Besacier,
  X.~Anguera, and E.~Dupoux, ``The zero resource speech challenge 2017,'' in
  \emph{Automatic Speech Recognition and Understanding (ASRU), 2017 IEEE
  Workshop on}.\hskip 1em plus 0.5em minus 0.4em\relax IEEE, 2017.

\bibitem{besacier2006towards}
L.~Besacier, B.~Zhou, and Y.~Gao, ``Towards speech translation of non written
  languages,'' in \emph{Spoken Language Technology Workshop, 2006. IEEE}.\hskip
  1em plus 0.5em minus 0.4em\relax IEEE, 2006, pp. 222--225.

\bibitem{bartels2016toward}
C.~Bartels, W.~Wang, V.~Mitra, C.~Richey, A.~Kathol, D.~Vergyri, H.~Bratt, and
  C.~Hung, ``Toward human-assisted lexical unit discovery without text
  resources,'' in \emph{Spoken Language Technology Workshop (SLT), 2016
  IEEE}.\hskip 1em plus 0.5em minus 0.4em\relax IEEE, 2016, pp. 64--70.

\bibitem{lignos2010recession}
C.~Lignos and C.~Yang, ``Recession segmentation: simpler online word
  segmentation using limited resources,'' in \emph{Proceedings of the
  fourteenth conference on computational natural language learning}.\hskip 1em
  plus 0.5em minus 0.4em\relax Association for Computational Linguistics, 2010,
  pp. 88--97.

\bibitem{godard2017very}
\BIBentryALTinterwordspacing
P.~Godard, G.~Adda, M.~Adda-Decker, J.~Benjumea, L.~Besacier,
  J.~Cooper-Leavitt, G.-N. Kouarata, L.~Lamel, H.~Maynard, M.~Mueller
  \emph{et~al.}, ``A very low resource language speech corpus for computational
  language documentation experiments,'' 2017, {arXiv}:1710.03501. [Online].
  Available: \url{http://arxiv.org/abs/1710.03501}
\BIBentrySTDinterwordspacing

\bibitem{hamlaoui2018}
F.~Hamlaoui, E.-M. Makasso, M.~Müller, J.~Engelmann, G.~Adda, A.~Waibel, and
  S.~Stüker, ``{BULBasaa}: A bilingual {B\`as\`a\'a-French} speech corpus for
  the evaluation of language documentation tools,'' in \emph{LREC 2018 (in
  press)}, Japan, 2018.

\bibitem{ludusan2014}
B.~Ludusan, M.~Versteegh, A.~Jansen, G.~Gravier, X.-N. Cao, M.~Johnson, and
  E.~Dupoux, ``Bridging the gap between speech technology and natural language
  processing: an evaluation toolbox for term discovery systems,'' in
  \emph{Proceedings of {LREC}}, 2014.

\bibitem{Goldwater09bayesian}
S.~Goldwater, T.~L. Griffiths, and M.~Johnson, ``A {Bayesian} framework for
  word segmentation: Exploring the effects of context,'' \emph{Cognition}, vol.
  112, no.~1, pp. 21--54, 2009.

\bibitem{aren}
A.~Jansen and B.~Van~Durme, ``Efficient spoken term discovery using randomized
  algorithms,'' in \emph{Proc. Automatic {Speech} {Recognition} and
  {Understanding} ({IEEE ASRU})}, 2011, pp. 401--406.

\bibitem{lee2015unsupervised}
C.-y. Lee, T.~J. O'Donnell, and J.~Glass, ``Unsupervised lexicon discovery from
  acoustic input,'' \emph{Transactions of the Association for Computational
  Linguistics}, vol.~3, pp. 389--403, 2015.

\bibitem{elsner2013joint}
M.~Elsner, S.~Goldwater, N.~Feldman, and F.~Wood, ``A joint learning model of
  word segmentation, lexical acquisition, and phonetic variability,'' in
  \emph{Proc. EMNLP}, 2013.

\bibitem{ney1999speech}
H.~Ney, ``Speech translation: Coupling of recognition and translation,'' in
  \emph{Proc. ICASSP}, vol.~1, 1999.

\bibitem{matusov2005integration}
E.~Matusov, S.~Kanthak, and H.~Ney, ``On the integration of speech recognition
  and statistical machine translation,'' in \emph{Ninth European Conference on
  Speech Communication and Technology}, 2005.

\bibitem{duong2016attentional}
L.~Duong, A.~Anastasopoulos, D.~Chiang, S.~Bird, and T.~Cohn, ``An attentional
  model for speech translation without transcription,'' in \emph{Proceedings of
  NAACL-HLT}, 2016, pp. 949--959.

\bibitem{berard2016listen}
\BIBentryALTinterwordspacing
A.~B{\'e}rard, O.~Pietquin, C.~Servan, and L.~Besacier, ``Listen and translate:
  A proof of concept for end-to-end speech-to-text translation,'' in
  \emph{Proc. NIPS Workshop on End-to-end Learning for Speech and Audio
  Processing}, 2016. [Online]. Available:
  \url{https://arxiv.org/abs/1612.01744}
\BIBentrySTDinterwordspacing

\bibitem{chan2016listen}
W.~Chan, N.~Jaitly, Q.~Le, and O.~Vinyals, ``Listen, attend and spell: A neural
  network for large vocabulary conversational speech recognition,'' in
  \emph{Proc. ICASSP}.\hskip 1em plus 0.5em minus 0.4em\relax IEEE, 2016, pp.
  4960--4964.

\bibitem{berard2018end}
A.~B{\'e}rard, L.~Besacier, A.~C. Kocabiyikoglu, and O.~Pietquin, ``End-to-end
  automatic speech translation of audiobooks,'' \emph{arXiv preprint
  arXiv:1802.04200}, 2018.

\bibitem{weiss2017sequence}
\BIBentryALTinterwordspacing
R.~J. Weiss, J.~Chorowski, N.~Jaitly, Y.~Wu, and Z.~Chen,
  ``Sequence-to-sequence models can directly transcribe foreign speech,'' in
  \emph{Proc. INTERSPEECH}, 2017. [Online]. Available:
  \url{https://arxiv.org/abs/1703.08581}
\BIBentrySTDinterwordspacing

\bibitem{anastasopoulos+chiang:naacl2018}
A.~Anastasopoulos and D.~Chiang, ``Tied multitask learning for neural speech
  translation,'' in \emph{Proc. NAACL HLT}, 2018, to appear.

\bibitem{bansal2017towards}
\BIBentryALTinterwordspacing
S.~Bansal, H.~Kamper, A.~Lopez, and S.~Goldwater, ``Towards speech-to-text
  translation without speech recognition,'' in \emph{Proc. EACL}, 2017.
  [Online]. Available: \url{http://aclweb.org/anthology/E17-2076}
\BIBentrySTDinterwordspacing

\bibitem{bansal2017weakly}
S.~Bansal, H.~Kamper, S.~Goldwater, and A.~Lopez, ``Weakly supervised spoken
  term discovery using cross-lingual side information,'' in \emph{Proc.
  ICASSP}.\hskip 1em plus 0.5em minus 0.4em\relax IEEE, 2017, pp. 5760--5764.

\bibitem{anastasopoulos2017spoken}
A.~Anastasopoulos, S.~Bansal, D.~Chiang, S.~Goldwater, and A.~Lopez, ``Spoken
  term discovery for language documentation using translations,'' in
  \emph{Proc. {Workshop on Speech-Centric Natural Language Processing}}, 2017,
  pp. 53--58.

\bibitem{kamper2017visually}
H.~Kamper, S.~Settle, G.~Shakhnarovich, and K.~Livescu, ``Visually grounded
  learning of keyword prediction from untranscribed speech,'' 2017,
  {arXiv}:1703.08136.

\bibitem{ip_stueker2008d}
S.~St{\"u}ker, ``Towards human translations guided language discovery for {ASR}
  systems,'' in \emph{Proc. {SLTU}}, Hanoi, Vietnam, May 2008.

\bibitem{stuker:hal-00959225}
S.~St\"uker, L.~Besacier, and A.~Waibel, ``{Human Translations Guided Language
  Discovery for ASR Systems},'' in \emph{Proc. {I}nterspeech}.\hskip 1em plus
  0.5em minus 0.4em\relax Brighton (UK): {Eurasip}, 2009, pp. 1--4.

\bibitem{StahlbergSVS12}
F.~Stahlberg, T.~Schlippe, S.~Vogel, and T.~Schultz, ``Word segmentation
  through cross-lingual word-to-phoneme alignment,'' in \emph{Spoken Language
  Technology Workshop (IEEE SLT)}, 2012, pp. 85--90.

\bibitem{boito2017unwritten}
M.~Z. Boito, A.~B{\'e}rard, A.~Villavicencio, and L.~Besacier, ``Unwritten
  languages demand attention too! word discovery with encoder-decoder models,''
  in \emph{Proc. IEEE ASRU}, 2017.

\bibitem{manolessou2005greek}
I.~Manolessou, ``The greek dialects of southern {I}taly: an overview,''
  \emph{KAMPOS: Cambridge Papers in Modern Greek}, vol.~13, pp. 103--125, 2005.

\bibitem{horrocks2009greek}
G.~Horrocks, \emph{Greek: A History of the Language and its Speakers}.\hskip
  1em plus 0.5em minus 0.4em\relax Wiley-Blackwell, 2009.

\bibitem{douri2015griko}
A.~Douri and D.~De~Santis, ``Griko and modern {G}reek in {G}rec{\`\i}a
  {S}alentina: an overview,'' \emph{L'Idomeneo}, vol. 2015, no.~19, pp.
  187--198, 2015.

\bibitem{chatzikyriakidis2010clitics}
S.~Chatzikyriakidis, ``Clitics in four dialects of modern {G}reek: A dynamic
  account,'' Ph.D. dissertation, University of London, 2010.

\bibitem{rohlfs1977grammatica}
G.~Rohlfs, \emph{Grammatica storica dei dialetti italogreci ({C}alabria,
  {S}alento) dt. Original [1949–1954] [Historical Grammar of the {I}taliot
  {G}reek dialects ({C}alabria, {S}alento)]}.\hskip 1em plus 0.5em minus
  0.4em\relax CH Beck, 1977.

\bibitem{karanastasis1997grammatiki}
A.~Karanastasis, \emph{Grammatiki ton ellinikon idiomaton tis {K}ato {I}talias
  [Grammar of the {G}reek dialects of south {Italy}]}.\hskip 1em plus 0.5em
  minus 0.4em\relax Akadimia Athinon, 1997.

\bibitem{anastasopoulos+al:coling2018}
A.~Anastasopoulos, M.~Lekakou, J.~Quer, E.~Zimianiti, J.~DeBenedetto, and
  D.~Chiang, ``Part-of-speech tagging on an endangered language: a parallel
  {Griko-Italian} resource,'' in \emph{Proc. COLING}, 2018, to appear.

\bibitem{palumbo1998io}
V.~D. Palumbo, \emph{Io' mia fora' - Fiabe e Racconti della {G}recìa
  {S}alentina [Once upon a time - Fairy Tales and Stories from {G}recìa
  {S}alentina]}.\hskip 1em plus 0.5em minus 0.4em\relax Calimera (LE):
  Gheton\'{i}a, 1998, a cura di S. Tommasi.

\bibitem{palumbo1999itela}
V.~D. Palumb{o}, \emph{'Itela na su p\'{o} - Canti popolari della Grecìa
  Salentina [I wanted to tell you - Folk songs of {G}rec\'{i}a
  {S}alentina]}.\hskip 1em plus 0.5em minus 0.4em\relax Calimera (LE):
  Gheton\'{i}a, 1999, a cura di S. Sicuro.

\bibitem{grikodatabase}
M.~Lekakou, V.~Baldiserra, and A.~Anastasopoulos, ``Documentation and analysis
  of an endangered language: aspects of the grammar of {Griko},'' 2013,
  \url{http://griko.project.uoi.gr}.

\bibitem{ondel2016variational}
L.~Ondel, L.~Burget, and J.~{\v{C}}ernock{\`y}, ``Variational inference for
  acoustic unit discovery,'' \emph{Procedia Computer Science}, vol.~81, pp.
  80--86, 2016.

\bibitem{anastasopoulos-chiang-duong:2016:EMNLP2016}
\BIBentryALTinterwordspacing
A.~Anastasopoulos, D.~Chiang, and L.~Duong, ``An unsupervised probability model
  for speech-to-translation alignment of low-resource languages,'' in
  \emph{Proc. EMNLP}, 2016. [Online]. Available:
  \url{https://aclweb.org/anthology/D16-1133}
\BIBentrySTDinterwordspacing

\bibitem{och2001statistical}
F.~J. Och and H.~Ney, ``Statistical multi-source translation,'' in
  \emph{Proceedings of MT Summit}, vol.~8, 2001, pp. 253--258.

\bibitem{petitjean2011global}
F.~Petitjean, A.~Ketterlin, and P.~Gan{\c{c}}arski, ``A global averaging method
  for dynamic time warping, with applications to clustering,'' \emph{Pattern
  Recognition}, vol.~44, no.~3, pp. 678--693, 2011.

\bibitem{goldwater2009bayesian}
S.~Goldwater, T.~L. Griffiths, and M.~Johnson, ``A bayesian framework for word
  segmentation: Exploring the effects of context,'' \emph{Cognition}, vol. 112,
  no.~1, pp. 21--54, 2009.

\bibitem{godard2016preliminary}
P.~Godard, G.~Adda, M.~Adda-Decker, A.~Allauzen, L.~Besacier,
  H.~Bonneau-Maynard, G.-N. Kouarata, K.~L{\"o}ser, A.~Rialland, and F.~Yvon,
  ``Preliminary experiments on unsupervised word discovery in mboshi,'' in
  \emph{Interspeech 2016}, 2016.

\end{thebibliography}

\end{document}